\documentclass[sigconf]{acmart}

\usepackage[inkscapelatex=false]{svg}
\usepackage{url}
\usepackage{hyperref}
\usepackage{graphicx}
\usepackage{booktabs}
\usepackage{multirow}
\usepackage{multicol}
\usepackage{xcolor}
\usepackage{colortbl}
\usepackage{amsmath}
\usepackage{amsfonts}
\usepackage{bbding}
\usepackage{enumitem}
\usepackage{subcaption}
\usepackage[most]{tcolorbox}
\usepackage[bottom,splitrule]{footmisc}

\renewcommand\footnotetextcopyrightpermission[1]{}

\AtBeginDocument{
  \providecommand\BibTeX{{
    \normalfont B\kern-0.5em{\scshape i\kern-0.25em b}\kern-0.8em\TeX}}}

\begin{document}

\title{TinyChart: Efficient Chart Understanding with \\Visual Token Merging and Program-of-Thoughts Learning}

\author{Liang Zhang}
\authornote{Authors contributed equally.}
\authornote{Work done during an internship at Alibaba Group.}
\affiliation{
  \institution{Renmin University of China}
  \city{Beijing}
  \country{China}
}
\email{zhangliang00@ruc.edu.cn}

\author{Anwen Hu}
\authornotemark[1]
\affiliation{
  \institution{Alibaba Group}
  \city{Beijing}
  \country{China}
}
\email{huanwen.haw@alibaba-inc.com}

\author{Haiyang Xu}
\affiliation{
  \institution{Alibaba Group}
  \city{Hangzhou}
  \country{China}
}
\email{shuofeng.xhy@alibaba-inc.com}

\author{Ming Yan}
\affiliation{
  \institution{Alibaba Group}
  \city{Hangzhou}
  \country{China}
}
\email{ym119608@alibaba-inc.com}

\author{Yichen Xu}
\affiliation{
  \institution{Renmin University of China}
  \city{Beijing}
  \country{China}
}
\email{xu_yichen@ruc.edu.cn}

\author{Qin Jin}
\authornote{Corresponding author.}
\affiliation{
  \institution{Renmin University of China}
  \city{Beijing}
  \country{China}
}
\email{qjin@ruc.edu.cn}

\author{Ji Zhang}
\affiliation{
  \institution{Alibaba Group}
  \city{Hangzhou}
  \country{China}
}
\email{zj122146@alibaba-inc.com}

\author{Fei Huang}
\affiliation{
  \institution{Alibaba Group}
  \city{Hangzhou}
  \country{China}
}
\email{f.huang@alibaba-inc.com}

\begin{abstract}
Charts are important for presenting and explaining complex data relationships. 
Recently, multimodal large language models (MLLMs) have shown remarkable capabilities in various chart understanding tasks. However, the sheer size of these models in terms of parameters and computational requirements limits their use in resource-constrained environments. In this paper, we present TinyChart, an efficient MLLM for chart understanding with only 3B parameters. TinyChart overcomes two key challenges in efficient chart understanding:  
(1) reduce the burden of learning numerical computations through a Program-of-Thoughts (PoT) learning strategy, which trains the model to generate Python programs for numerical calculations, and
(2) reduce lengthy vision feature sequences produced by the vision transformer for high-resolution images through a Vision Token Merging module, which gradually merges most similar vision tokens.
Extensive experiments demonstrate that our 3B TinyChart achieves SOTA performance on a variety of chart understanding benchmarks including ChartQA, Chart-to-Text, Chart-to-Table, OpenCQA, and ChartX. It outperforms several chart understanding MLLM with up to 13B parameters such as ChartLlama and ChartAst, and close-sourced general-purpose MLLM GPT-4V on ChartQA. It also demonstrates its superior efficiency with higher throughput during inference due to a smaller model scale and more efficient vision encoding. Our code and model are available at \href{https://github.com/X-PLUG/mPLUG-DocOwl/tree/main/TinyChart}{https://github.com/X-PLUG/mPLUG-DocOwl/tree/main/TinyChart}.
\end{abstract}

\maketitle

\section{Introduction}
\begin{figure}[t]
\includegraphics[width=0.9\linewidth]{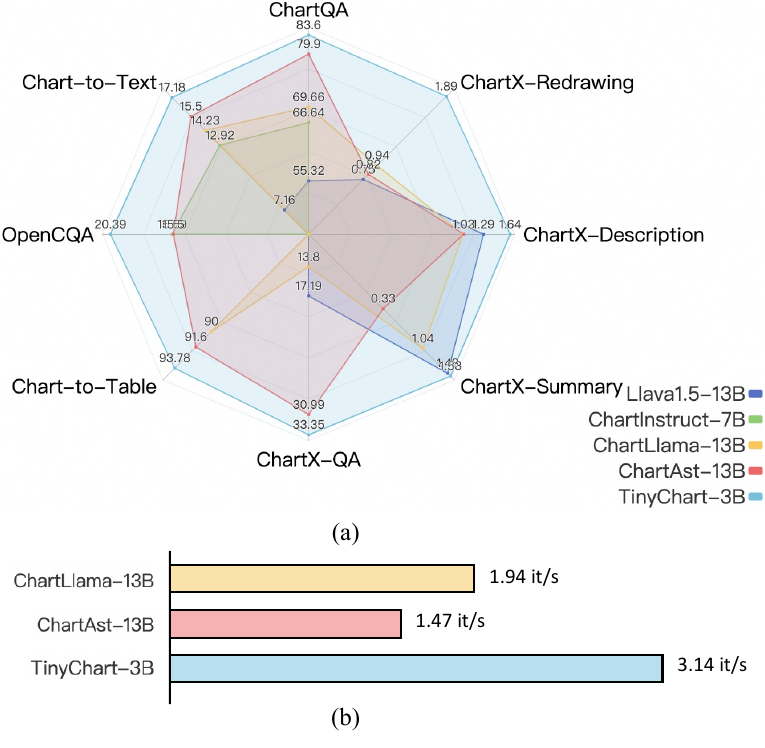}
\vspace{-4pt}
\caption{Our TinyChart-3B outperforms several 13B MLLMs on a variety of chart understanding benchmarks (a), while achieving larger inference throughput (b).}
\label{fig:radar}
\vspace{-8pt}
\end{figure}
 
As an important information source, charts can intuitively visualize data in various visual presentation forms and have become an indispensable part of information dissemination, business decision-making, and academic research~\cite{chartsurvey}. With the rapid growth of multimodal data, automatically comprehending charts has become a pressing need and received increasing attention from the research community~\cite{chartllama,chartast,chartinstruct,onechart}. Recently, Multimodal Large Language Models (MLLMs) have shown strong capability in comprehending images and following instructions~\cite{gpt4,llava,mplugowl,llava1.5,sphinx,mplugowl2, xcomposer, xcomposer2, xcomposer2-4k}. Based on these MLLMs, some recent works~\cite{chartllama,chartast,chartinstruct,paperowl} further build chart understanding models by collecting and constructing versatile chart comprehension datasets and performing supervised fine-tuning. 

However, despite their remarkable success, current chart understanding models still face three main limitations: (1) Considerable amount of parameters makes training and deployment challenging. For example, ChartLlama~\cite{chartllama} is a model with 13 billion parameters, which is hard to deploy on a single consumer GPU with less than 26GB of VRAMs.
(2) They are prone to errors when tackling questions involving numerical calculations~\cite{chartast}, which are difficult to directly answer without any reasoning steps.
(3) They struggle with efficiently encoding for high-resolution images since the standard vision transformer would produce lengthy feature sequences. 

To overcome such limitations in chart understanding, we propose an efficient and powerful MLLM, namely~\textbf{TinyChart}. As shown in Figure~\ref{fig:radar}, through the efficient visual encoding and Program-of-Thoughts learning strategy, TinyChart achieves state-of-the-art performances on various chart understanding benchmarks with only 3B parameters, while excelling in faster inference throughput.

For efficient visual encoding, we propose to merge visual tokens based on the observation that chart images often contain large areas of color and white spaces.
Inspired by~\cite{tome}, we adopt a parameter-free Visual Token Merging module inside each vision transformer layer, which aggregates the most similar visual tokens and gradually reduces the length of the visual feature sequence, thus making it possible to efficiently encode high-resolution chart images. 
This enables the model to maintain high-resolution chart image input while controlling the computation load. 

Moreover, inspired by ~\cite{pot}, we propose Program-of-Thoughts learning that enhances the model's ability to resolve mathematical problems. According to statistics on ChartQA~\cite{chartqa}, 42\% of questions for charts require numerical answers, and most existing models struggle to perform numerical question answering~\cite{matcha, chartast}. To learn chart understanding more efficiently, we train the model to generate Python programs for the computation problems step by step. The programs are then passed to a Python interpreter to produce the final answer. To support Program-of-Thoughts learning, we further construct the ChartQA-PoT dataset based on ChartQA~\cite{chartqa}. The QA pairs in our ChartQA-PoT are constructed in two ways: (1) Template-based PoT construction, which generates questions and programs by filling in the manually written templates based on chart data. (2) GPT-based PoT construction, which leverages \texttt{gpt-3.5-turbo}~\cite{gpt3.5} to generate programs based on human-written questions. Experimental results show that Program-of-Thoughts learning can significantly improve the question-answering, especially numerical question answering ability of TinyChart.\\ 
The main contributions of this work are as follows:
\begin{itemize}[leftmargin=*]
\item We introduce TinyChart, an efficient multimodal chart understanding model, which outperforms several 13B MLLMs and achieves state-of-the-art performances on a variety of chart understanding benchmarks, while excelling in faster inference speed at the same time.
\item We propose a Program-of-Thoughts (PoT) learning strategy to enhance the model in learning numerical calculation and carefully build a PoT dataset ChartQA-PoT. 
\item We adopt Visual Token Merging for efficient vision encoding, which significantly reduces the length of vision feature sequences and enables the model to encode high-resolution chart images with constrained computing resources.
\end{itemize}
\section{Related Work}
\subsection{Chart Understanding}
Chart understanding requires the model to comprehend chart contents and accomplish related tasks specified by the instructions. This field encompasses low-level recognition tasks, such as data extraction~\cite{deplot}, and high-level tasks, such as question-answering  (QA)~\cite{chartqa,plotqa,dvqa}, summarization~\cite{chart2text,chart2text-8k}, and re-rendering~\cite{chartllama}. As charts often contain OCR text pivotal for data interpretation, and many instructions require the model to perform numerical calculations, chart understanding demands robust text recognition capabilities and computational reasoning from the model. Early approaches~\cite{lorra, plotqa, deplot,chartstamp, mpmqa, qc_cap} rely on pipeline methods that use off-the-shelf OCR tools or component detectors to transform charts into data tables and other textual representations. They then employ language models to complete the specified tasks. These pipeline approaches, limited by their inability to optimize jointly, were hampered by error accumulation. Recent studies~\cite{unichart, matcha, chartllama, chartast,chartinstruct,mmc} have shifted towards end-to-end methods based on multimodal large language models. These studies adopt the structure of multimodal large language models~\cite{llava,llava1.5,mplugowl,mplugowl2,sphinx} and enhance chart understanding abilities through supervised fine-tuning~\cite{instructgpt} with substantial chart instruction data~\cite{chartllama,chartast,chartinstruct}. 
Although these models demonstrate improvement in performance, their extensive parameter size prevents them from being easily trained or deployed under resource-constrained scenarios. In this paper, we demonstrate that a 3B MLLM is enough to achieve state-of-the-art performance on several chart understanding tasks.
Meanwhile, it has been well observed that these models are prone to numerical errors~\cite{matcha,chartinstruct,chartast}. 
Though~\citet{chartast} try to construct executable command lines in JSON format based on a template to eliminate numerical errors, we argue that it is insufficient to fully address this issue for two reasons: 1) The executable command lines in JSON format produced by~\citet{chartast} relies on a specific computational backend, which limits their potential versatility. 2) Template-based programs can only cover rather limited scenarios. Instead, we construct the Program-of-Thoughts learning dataset with the combination of both templates and GPT-generated programs. 
This allows the model to more effectively learn how to solve numerical problems.
\subsection{Multimodal Large Language Model}
Multimodal large language models (MLLM) exhibit strong capabilities in visual understanding and instruction following~\cite{gpt4,gemini}. They typically comprise transformer-based visual encoders, large language models, and vision-language connectors~\cite{llava,llava1.5,tinyllava,mplugowl,mplugowl2,xcomposer,xcomposer2,mplug-octopus}. These models are generally trained on extensive general image-text data for cross-modal alignment and instruction fine-tuning. Although some studies have showcased a degree of OCR capability in these multimodal large language models~\cite{ocr_mllm,trie}, their performance on document and chart understanding benchmarks remains suboptimal due to their low input resolution~\cite{ureader,xcomposer2-4k}. Efforts in the general document domain have attempted to improve the fine-grained understanding capabilities of MLLMs by increasing resolution~\cite{qwenvl}, segmenting images~\cite{ureader,sphinx,docowl1.5,xcomposer2-4k}, utilizing frequency domain signals~\cite{docpedia}, and introducing additional high-resolution encoders~\cite{cogagent}. However, these models often suffer from low efficiency, primarily due to the excessive length of the high-resolution visual sequences. The visual token merging method adopted in this paper can significantly reduce the length of visual feature sequences and relax the computational requirements with high-resolution input.
\begin{figure*}[h]
\includegraphics[width=0.9\linewidth]{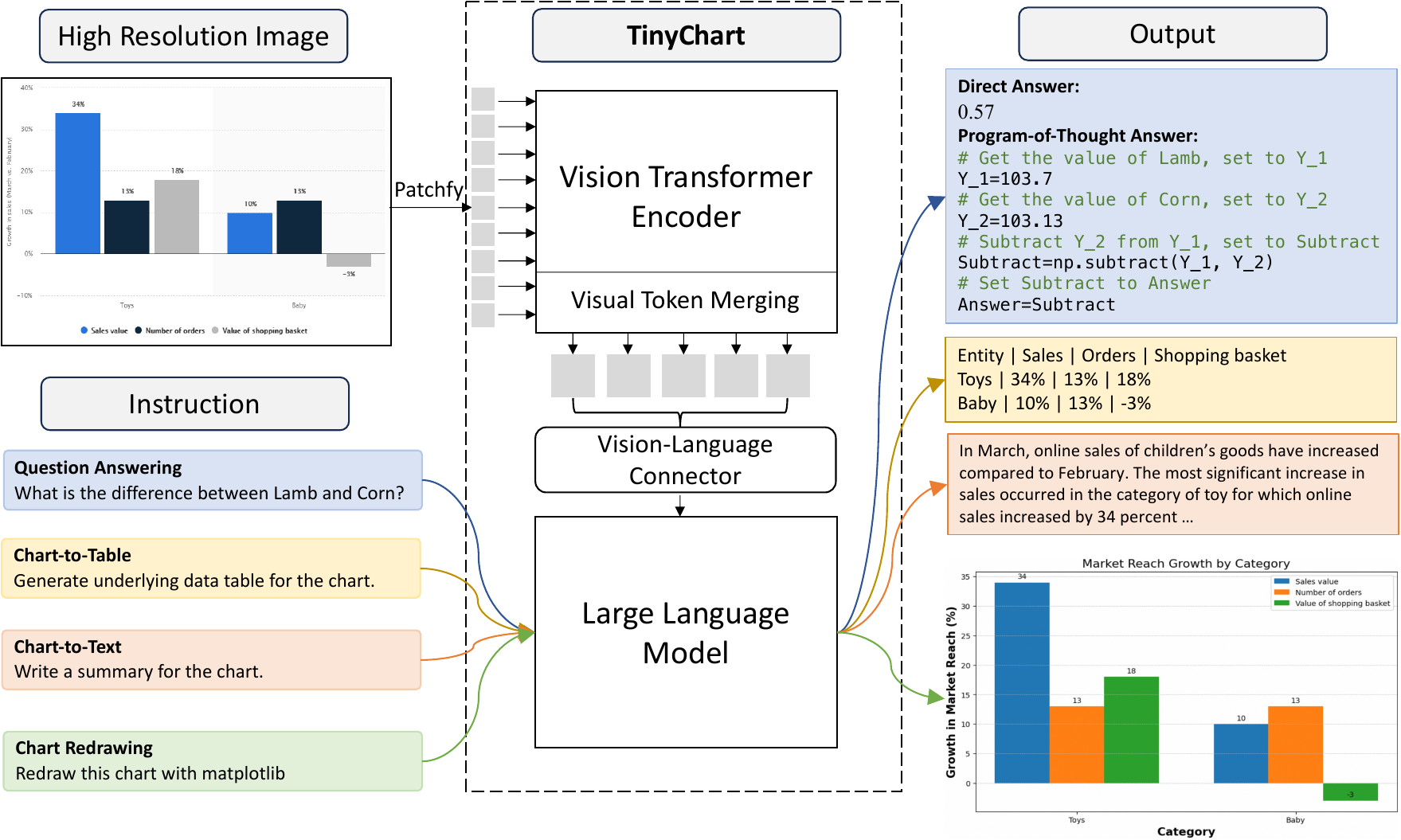}
\vspace{-8pt}
\caption{Overview of TinyChart.}
\label{fig:overview}
\end{figure*}

\begin{figure*}[ht]
\includegraphics[width=0.9\linewidth]{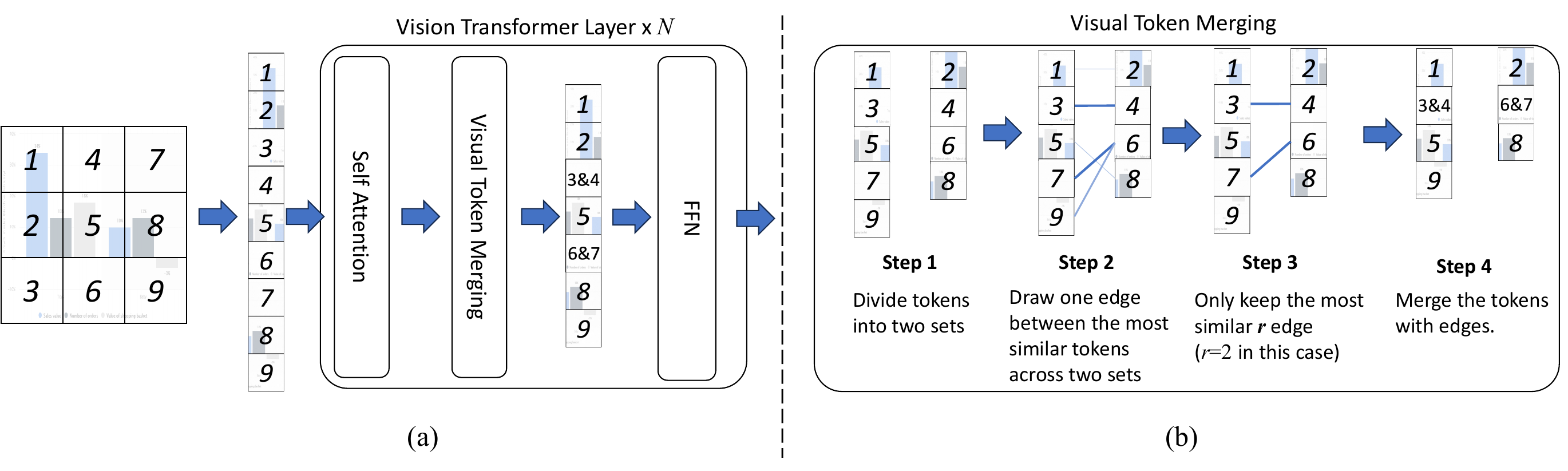}
\vspace{-8pt}
\caption{(a) Vision transformer layer with Visual Token Merging. (b) Process of the Visual Token Merging.}
\label{fig:tokenmerge}
\end{figure*}

\section{TinyChart}
\subsection{Model Architecture}
Figure~\ref{fig:overview} shows the overview framework of our proposed TinyChart. It follows the typical architecture of the multimodal large language model (MLLM), which consists of a vision transformer encoder, a vision-language connector, and a large language model. To encode high-resolution visual input effectively, we insert the visual token merging module inside each vision transformer layer. 
\subsubsection{Vision Transformer Encoder}
The vision transformer encoder aims to encode chart images into vision features. A standard vision transformer~\cite{vit} first resizes the input image \(I\) into a fixed resolution and crops the image into patches. Then the patches are treated as vision tokens and processed with transformer encoder layers~\cite{transformer}. Suppose the input image \(I^{N\times N}\) is in resolution \(N \times N\), and the patch size is \(P \times P\), the length of vision tokens would be \((N // P)^2\). Since the standard transformer layer does not reduce the sequence length, the vision transformer finally produces a vision feature in length \((N // P)^2\). In practice, when \(N\) is large, the vision feature can be very long and inefficient for the language model to handle. 

\begin{figure*}[h]
\includegraphics[width=0.9\linewidth]{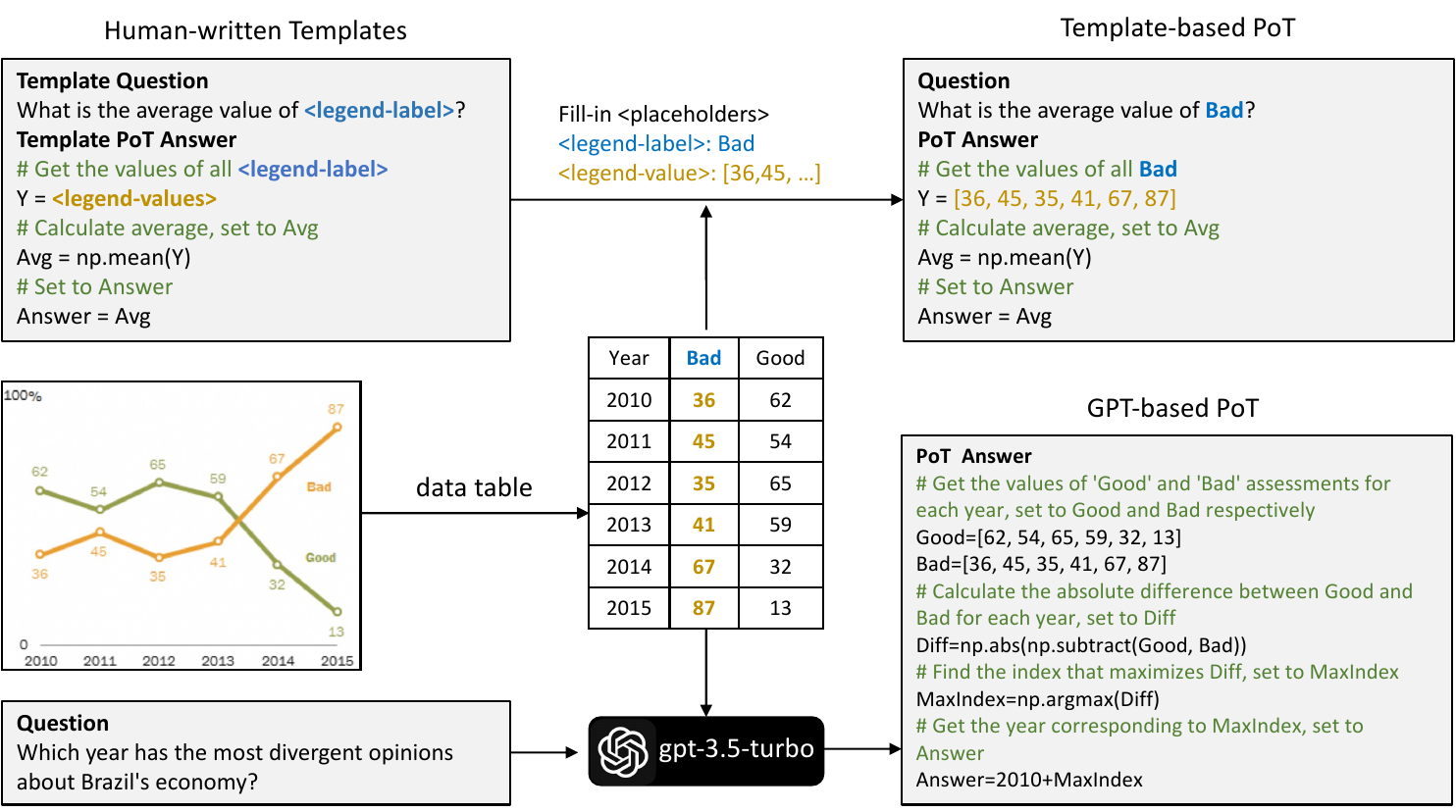}
\caption{The demonstration of constructing  Template-based PoT (upper half) and GPT-based PoT (lower half) in the ChartQA-PoT dataset.}
\label{fig:pot_construction}
\end{figure*}

\noindentparagraph{\textbf{Visual Token Merging}}
Since key information (such as OCR words) in a chart can be unrecognizable in low-resolution images~\cite{docowl1.5}, high-resolution input is essential for chart understanding. However, charts typically contain a large number of color blocks and blank spaces, where patches are visually similar.
To achieve efficient and effective chart understanding, we apply Visual Token Merging~\cite{tome} in each transformer layer. The process of Visual Token Merging is shown in Figure~\ref{fig:tokenmerge}. By merging the \(r\) most similar token pairs, it reduces the length of the vision feature by \(r\) in each layer. We measure the similarity between two tokens using the cosine distance between Keys from self-attention following~\cite{tome}. As shown in the lower part of Figure~\ref{fig:tokenmerge}, Vision Token Merger finds the top-\(r\) similar token pairs through bipartite graph matching. It first divides the vision tokens into two disjoint sets. Then, for each token in one set, it finds the most similar tokens in the other set and draws an edge between the two tokens. After that, it only keeps the top-\(r\) most similar edges and merges the features of the two endpoints through average pooling. Note that not only spatially adjacent visual tokens are subject to merging. 
Non-adjacent tokens can also be merged if they belong to different subsets and are similar enough.

\noindentparagraph{\textbf{Proportional attention}} The visual token merging operation aggregates tokens with a similar feature into one. Therefore, it will reduce the proportion of this visual feature in the attention calculation in the following transformer layer, since the number of this feature has decreased. To solve this issue, we let the attention operation consider the actual number of patches \(s\) represented by each token as follows:
\begin{align}
\mathrm{Attention}=\mathrm{softmax}\left( \frac{QK^\top}{\sqrt{d}} + \log s \right) V
\end{align}

Where \(Q\), \(K\), \(V\) denotes the query, key, and value of self-attention which are linear projected from the hidden states~\cite{transformer}. By adding \(\log s\) inside \(\mathrm{softmax}\), the token that merged from \(s\) patches are duplicated by \(s\) times in the attention calculation~\cite{tome}.

\subsubsection{Vision-Language Connector} The vision language connector aims to project the vision features into the embedding space of the large language model. Following~\cite{llava1.5, tinyllava}, we implement the vision-language connector as a multiple-layer perceptron with one hidden layer and GeLU~\cite{gelu} activation.

\subsubsection{Large Language Model} The large language model aims to comprehend both visual features and language instructions, and then generate responses to accomplish chart understanding tasks. It is implemented as a transformer decoder~\cite{transformer} with a causal attention mask. The training objective of the model is language modeling. Assuming the visual features is \(V\), the language instruction is \(L\), and the response is \(R\), then the loss function is defined as follows:
\begin{align}
\mathcal{L}=\frac{1}{T}\sum_{i=1}^T\mathrm{LLM}(R_i|V, L, R_{<i})
\label{eq:loss}
\end{align}
Where \(T\) is the number of tokens in \(R\). Note that we only calculate loss over tokens in the responses following the supervised fine-tuning setting in~\cite{llava1.5}.

\subsection{Program-of-Thoughts Learning}
Program-of-Thoughts (PoT) learning aims to enhance the learning efficiency of models for numerical computation. 
In PoT learning, the model is trained to generate executable Python codes as the target of a question. The final answer is obtained by executing the code with a Python interpreter. 
Compared to short answers that only contain the calculated values, the Python code includes natural language comments and multi-step reasoning processes, offering a form of learning closely aligned with the pre-training of the large language model.

\noindentparagraph{\textbf{ChartQA-PoT Dataset}} To support PoT learning on chart understanding, we construct the ChartQA-PoT dataset based on the training split of ChartQA~\cite{chartqa}. ChartQA-PoT contains 140,584 (question, PoT answer) pairs. Each PoT answer consists of multiple lines of Python code. We provide natural language comments for almost all code lines to explain their behaviors.
We employ two approaches for constructing (question, PoT answer) pairs: Template-based  PoT, and GPT-based PoT. 

\subsubsection{Template-based PoT}Based on the chart images in ChartQA, we construct template-based (question, PoT answer) pairs. As illustrated in the upper half of Figure~\ref{fig:pot_construction}, the Template-based PoT is constructed based on human-written templates containing placeholders for both questions and code. The template questions involve common numerical operations such as calculating the sum, average, minimal, and maximum values. We adopt the 40 template questions proposed by PlotQA~\cite{plotqa} and manually write their corresponding template Python code to solve them. As shown in the top-left part of Figure~\ref{fig:pot_construction}, the template code consists of several variable assignment operations with NumPy~\cite{numpy} functions to perform calculations. The beginning steps usually involve extracting the relevant data from the chart and assigning them to variables. The final computed result is stored in a variable named "Answer". For each placeholder in the template, we identify all possible values from the data table of the chart and randomly select one to fill in the placeholder. After removing incorrect or unreasonable filled-ins using rule-based methods, we finally successfully construct 119,281 (question, PoT pairs) over 17,498 images from ChartQA. 

\subsubsection{GPT-based PoT}
Although the template-based method allows for the construction of a large number of question-answer pairs, the diversity of these pairs is limited due to the fixed templates. To improve the generalization ability of PoT learning, we have additionally built GPT-generated PoT data by leveraging the powerful command-following and code-generation capabilities of large language models. Specifically, we prompt \texttt{gpt-3.5-turbo}~\cite{gpt3.5} to generate PoT answers similar to the template PoT format for questions annotated in ChartQA using in-context examples. As shown in Figure~\ref{fig:pot_construction}, since \texttt{gpt-3.5-turbo} does not accept image input, we also provide the data table corresponding to the chart as text input to \texttt{gpt-3.5-turbo}. We screen the quality of the generated PoT answers by running them through a Python interpreter. If the annotated PoT answer can not run on the Python interpreter, or if the answer obtained is different from the annotated one in ChartQA, then the corresponding PoT Answer is deleted. In the end, we construct 21,303 (question, PoT Answer) pairs on 15,521 chart images.

\begin{table}[h!]
\caption{Datasets used for training TinyChart. The benchmark datasets consist of basic chart understanding evaluations including QA, summary, and chart-to-table generation. Note that in ablation studies, we only use the benchmark datasets for training due to limited computational resources. }
\vspace{-8pt}
\begin{tabular}{@{}lcr@{}}
\toprule
Dataset & Benchmark & Samples \\ \midrule
\textbf{\textit{Chart question answer}} &  &  \\
ChartQA~\cite{chartqa} & \checkmark & 28,299 \\
ChartQA-PoT & \checkmark & 140,584 \\
PlotQA~\cite{plotqa} &  & 157,070 \\
DVQA~\cite{dvqa} &  & 200,000 \\
OpenCQA~\cite{opencqa} & & 5,407 \\ \midrule
\textbf{\emph{Chart-to-text generation}} &  &  \\
Pew~\cite{chart2text} & \checkmark & 7,892 \\
Statista~\cite{chart2text} & \checkmark & 29,589 \\
OpenCQA~\cite{opencqa} &  & 5,407 \\
Vistext~\cite{vistext} &  & 11,171 \\
ChartSumm~\cite{chartsumm} &  & 75,255 \\
Chart2Text-8k~\cite{chart2text-8k} &  & 7,862 \\ \midrule
\textbf{\emph{Chart-to-table generation}} &  &  \\
ChartQA~\cite{chartqa} & \checkmark & 19,373 \\
PlotQA~\cite{plotqa} &  & 190,720 \\
Chart2Text-8k &  & 8,305 \\
DVQA~\cite{dvqa} &  & 300,000 \\
Statista~\cite{chart2text} &  & 29,589 \\ \midrule
\textbf{\emph{Chart instruction following}} &  &  \\
ChartLlama~\cite{chartllama} &  & 148,398 \\ \midrule
\textbf{Total} &  & \textbf{1,364,921} \\ \bottomrule
\end{tabular}

\label{tab:training_data}
\end{table}

\subsection{Multitask Learning}
We perform multitask learning to train our TinyChart model. We collect a chart understanding dataset that contains 1.36M samples for supervised fine-tuning. It covers various chart understanding tasks including chart question answering, chart-to-text generation, chart-to-table generation, and chart instruction following.  Table~\ref{tab:training_data} shows the collection of our training dataset.
We mix data in different tasks together to jointly train the model, and use task-specified instructions to enable the model to differentiate between them. The training objective is language modeling on response tokens as presented in Eq.\ref{eq:loss}. Note that in ablation studies, we train solely with benchmark datasets due to limited computational resources.
\begin{table*}[ht]
\caption{Main results on chart-related benchmarks. The inference throughput is evaluated on the ChartQA test with a batch size of 1 on V100 32GB.}
\vspace{-8pt}
\label{tab:main_result}
\begin{tabular}{@{}lrcccccccc@{}}
\toprule
\multirow{2}{*}{Model} & \multirow{2}{*}{\#Parameters} & \multirow{2}{*}{Resolution} & \multirow{2}{*}{\begin{tabular}[c]{@{}c@{}}Inference\\ Throughput\end{tabular}} & \multicolumn{3}{c}{ChartQA} & Chart-to-Text & Chart-to-Table & OpenCQA \\ \cmidrule(l){5-10} 
 &  &  &  & Aug. & Hum. & Avg. & BLEU4 & RMS\(_{F1}\) & BLEU4 \\ \midrule
\multicolumn{2}{l}{\textbf{\emph{Close source models}}} \\
GPT-4V~\cite{gpt4} & - & - & - & - & - & 78.50 & - & - & -\\
Gemini-Ultra~\cite{gemini} & - & - & - & - & - & 80.80 & - & - & -\\
Qwen-VL-Max~\cite{qwenvl} & - & - & - & - & - & 79.80 & - & - & - \\ 
Deplot+Codex~\cite{deplot} & 1.3B+175B & - & - & 91.00 & 67.60 & 79.30 & - & 87.22 & - \\
\midrule
\multicolumn{2}{l}{\textbf{\emph{Open source models}}} \\
Llava1.5~\cite{llava1.5} & 13B & 336\(\times\)336 & 1.94 it/s & 72.96 & 37.68 & 55.32 & 7.16 & 48.95 & - \\
Qwen-VL~\cite{qwenvl} & 9.6B & 448\(\times\)448 & 1.65 it/s & 78.90 & 44.30 & 61.60 & - & - & - \\
UReader~\cite{ureader} & 7B & 224\(\times\)224(\(\times\)20) & 1.67 it/s & 79.42 & 39.12 & 59.30 & - & - & -\\
DocOwl1.5~\cite{docowl1.5} & 8B & 448\(\times\)448(\(\times\)9) & 1.56 it/s & 91.38 & 49.62 & 70.50 & - & - & -\\
ChartInstruct~\cite{chartinstruct} & 7B & - & - & 87.76 & 45.52 & 66.64 & 13.83 & - & 15.59 \\
ChartLlama~\cite{chartllama} & 13B & 336\(\times\)336 & 1.94 it/s & 90.36 & 48.96 & 69.66 & 14.23 & 90.00 & - \\
ChartAst~\cite{chartast} & 13B & 448\(\times\)448 & 1.47 it/s & \textbf{93.90} & 65.90 & 79.90 & 15.50 & 91.60 & 15.50 \\ \midrule
TinyChart@512 & 3B & 512\(\times\)512 & \textbf{3.65} it/s & 93.60 & 72.16 & 82.88 & \textbf{17.93} & 92.93 & 19.62 \\ 
TinyChart@768 & 3B & 768\(\times\)768 & 3.14 it/s & 93.86 & \textbf{73.34} & \textbf{83.60} & 17.18 & \textbf{93.78} & \textbf{20.39} \\ \bottomrule
\end{tabular}
\end{table*}

\section{Experiment}
\subsection{Implementation Details}
TinyChart is initialized from TinyLlava~\cite{tinyllava}, which utilizes the SigLIP~\cite{siglip} as the vision encoder and Phi-2~\cite{phi1.5} as the large language model. The origin input resolution of the vision encoder is 384\(\times\)384. We extend the input resolution to 512\(\times\)512 and 768\(\times\)768 and apply visual token merging with \(r=20\) and \(r=84\) in each transformer layer respectively. We train the entire model for 3 epochs with a batch size of 512. The learning rate is set to \(1e-4\), with a warmup in the beginning 3\% steps, and then decays to 0 at the end of training. The total training process costs 3 days on 32 Tesla V100 GPUs with 32 GB VRAMs.

\subsection{Evaluation Benchmarks}
\noindentparagraph{\textbf{ChartQA}} ChartQA~\cite{chartqa} aims to generate a short answer to the question based on the chart content. It includes a lot of questions that require numerical calculation.
We report the relaxed accuracy that allows numerical error within 5\% as the metric following~\cite{chartqa,chartllama,chartast}. Note that our TinyChart with Program-of-Thoughts learning can perform ChartQA in the following four settings:
\begin{itemize}[leftmargin=*]
\item \textbf{Direct}: the model produces short answers directly.
\item \textbf{PoT}: the model produces Python code. The answer is then calculated through the Python interpreter.
\item \textbf{Combine}: the model produces Python code for questions that require calculation, and Direct answers for others. We determine whether a question requires calculation with a simple rule-based keyword detector. If the question contains one of the calculative keywords\footnote{sum, mean, average, ratio, mode, divide, dividing, differ, subtract, add, division, times, absolute, minus, exceed, below, less, fewer, bigger, biggest, greater, higher, longer, tallest, lowest, number, how many colors, what is the value}, the detector will treat it as a computational question and prompt the model to generate a PoT answer. Otherwise, the model is instructed to produce a Direct answer. Additionally, if the generated program of a calculative question encounters syntax errors, we let the model produce Direct answers for this question in the Combine setting.
\item \textbf{Oracle}
We further introduce the Oracle setting for ChartQA evaluation. Under this setting, we always choose the correct one between the Direct and PoT answers after evaluating under both settings. It is the upper bound of the combination across the two answers.

\end{itemize} 
We evaluate TinyChart under the Combine setting by default.

\noindentparagraph{\textbf{Chart-to-Text}} Chart-to-Text aims to generate a chart summarization based on chart content. We evaluate the model with the Pew benchmark~\cite{chart2text}, and report BLEU4~\cite{bleu} as the metric.

\noindentparagraph{\textbf{Chart-to-Table}} Chart-to-Table aims to extract the underlying data table presented by the chart. We evaluate the performance of Chart-to-Table with the data table annotation provided by ChartQA~\cite{chartqa} following~\cite{chartllama, chartast}. We report RMS\(_{F1}\)~\cite{deplot} as the metric.

\noindentparagraph{\textbf{OpenCQA}} Different from ChartQA, OpenCQA~\cite{opencqa} evaluates the ability of models to generate free-form answers to the chart-related questions. We report BLEU4~\cite{bleu} as the metric following~\cite{chartinstruct, chartast}.

\noindentparagraph{\textbf{ChartX}} ChartX~\cite{chartx} is a recently proposed benchmark that contains more chart types. We evaluate the ChartX cognition tasks since they are more challenging. It covers Question Answering, Chart Description Generation, Chart Summary Generation, and Chart Redrawing. We report the GPT-Accuracy for QA and GPT-score for the remaining 3 tasks as the metrics following ChartX~\cite{chartx}.

\subsection{Main Results}
Table~\ref{tab:main_result} shows an extensive comparison between TinyChart and existing multimodal large language models on 4 chart understanding benchmarks. Our TinyChart model achieves state-of-the-art performance on ChartQA, Chart-to-Text, Chart-to-Table, and OpenCQA, while excels in larger inference throughput. Specifically, with the input resolution set at 768\(\times\)768, TinyChart achieves an accuracy of 83.60 on ChartQA~\cite{chartqa}, surpassing several closed-source models including GPT-4V, Gemini-Ultra, and Qwen-VL-Max~\cite{qwenvl}. It also outperforms previous open-source SOTA ChartAst~\cite{chartast} on chart understanding. 

We find that previous models performed poorly on the ChartQA human subset, with none of them achieving over 70\%. In contrast, the performance on the ChartQA-augmentation has approached 93.9\%. This is because the questions posed by human annotators involve more computational problems~\cite{chartqa} and are more challenging. By leveraging the Program-of-Thoughts learning, TinyChart achieves performance of 73.34\% on ChartQA-human, which is an improvement of 7.44\% over the previous state-of-the-art ChartAst~\cite{chartast}. This demonstrates the effectiveness of our proposed learning method based on the Program-of-Thoughts. 

We observed that models with higher input resolutions generally perform better on chart understanding tasks. However, encoding high-resolution charts leads to a decrease in inference speed (e.g., Qwen-VL vs. Llava1.5, DocOwl1.5 vs. UReader, ChartAst vs. ChartLlama). By leveraging visual token merging, TinyChart is able to accept higher-resolution input images with a limited increase in computing demands, thus achieving better performance. Due to the smaller model size and the efficient visual token merging strategy, TinyChart achieves significantly larger inference throughput compared to previous models. In summary, these results demonstrate that TinyChart can achieve efficient chart understanding with enhanced performance and faster inference.

\noindentparagraph{\textbf{ChartQA performance under different settings.}}Table~\ref{tab:chartqa_setting} shows the performance comparison under different settings. Note that the performance of ChartAst under the Combine setting is from~\citet{chartast}, which leverages a combination of Direct answer and executive JSON to produce the final answer. The results indicate that our TinyChart model could achieve SOTA performance on the Direct answer. By combining with PoT answers, TinyChart could make further improvements. In addition, since the combination of Direct and PoT answers is very simple, the performance under the Combine setting falls behind the Oracle setting a lot. Further study can be conducted to better combine the two answers.

\noindentparagraph{\textbf{Calculative and non-calculative questions.}} We divide the questions in ChartQA test set~\cite{chartqa} into two categories: calculative questions (761 of 2500) and non-calculative questions (1739 of 2500) by checking whether they contain calculative keywords mentioned above. Table~\ref{tab:cal_questions} shows the performance of TinyChart@768 on these two types of questions under different settings. We observe that PoT significantly improves the performance on calculative questions compared to Direct settings (78.98 vs. 56.64) and thus it shows overall performance gains (80.84 vs. 76.36). And the simple combination of Direct and PoT strategies further makes improvements.

\begin{table}[t]
\caption{Performance on ChartQA under different settings.}
\label{tab:chartqa_setting}
\begin{tabular}{@{}lcccc@{}}
\toprule
\multirow{2}{*}{Model} & \multicolumn{4}{c}{ChartQA} \\ \cmidrule(l){2-5} 
 & Direct & PoT & Oracle & Combine \\ \midrule
ChartLlama~\cite{chartllama} & 69.66 & - & - & - \\
ChartAst~\cite{chartast} & 75.10 & - & - & 79.90 \\ \midrule
TinyChart@512 & \textbf{76.92} & 79.64 & 88.76 & 82.88 \\
TinyChart@768 & 76.36 & \textbf{80.84} & \textbf{89.12} & \textbf{83.60} \\ \bottomrule
\end{tabular}
\end{table}

\begin{table}[]
\caption{ChartQA performance on Calculative (Cal.) and Non-calculative (Non-cal.) questions.}
\label{tab:cal_questions}
\scalebox{0.9}{
\begin{tabular}{@{}llccc@{}}
\toprule
\multirow{2}{*}{Model} & \multirow{2}{*}{Setting} & \multicolumn{3}{c}{ChartQA} \\ \cmidrule(l){3-5} 
& & Cal. (761) & Non-cal. (1739) & Total (2500) \\ \midrule
TinyChart@768 & Direct & 56.64 & \textbf{84.99} & 76.36 \\
TinyChart@768 & PoT & 78.98 & 81.66 & 80.84 \\
TinyChart@768 & Combine & \textbf{80.42} & \textbf{84.99} & \textbf{83.60} \\ \bottomrule
\end{tabular}
}
\end{table}

\noindentparagraph{\textbf{Evaluation on ChartX.}}To further assess the generalizability of TinyChart, we compare our model with end-to-end General MLLM and Chart MLLM on ChartX-Cognition benchmark~\cite{chartx}, since it covers visually diverse chart types. We use TinyChart@768 to perform inference on ChartX without additional fine-tuning. As shown in Table~\ref{tab:chartx}, benefiting from our Program-of-Thoughts learning method, TinyChart achieves a 33.35 GPT-Accuracy on the QA task, even surpassing the GPT-4V model. Though it falls behind GPT-4V in Summary, Description, and Redrawing tasks, TinyChart still performs better than open-source Chart MLLMs including ChartLlama and ChartAst. It indicates that TinyChart has a strong capability to generalize across various chart types.

\begin{table}[]
\caption{Evaluation results on ChartX~\cite{chartx}.}
\label{tab:chartx}
\begin{tabular}{@{}lcccc@{}}
\toprule
\multirow{2}{*}{Model} & \multicolumn{4}{c}{ChartX Cognition} \\ \cmidrule(l){2-5} 
 & QA & Summary & Description & Redrawing \\ \midrule
\multicolumn{2}{l}{\textbf{\emph{General MLLM}}} &  &  &  \\
Llava1.5 & 17.19 & 1.48 & 1.29 & 0.75 \\
GPT-4V & 33.04 & \textbf{3.17} & \textbf{3.12} & \textbf{2.63} \\ \midrule
\multicolumn{2}{l}{\textbf{\emph{Chart MLLM}}} &  &  &  \\
ChartLlama & 13.80 & 1.04 & 1.02 & 0.94 \\
ChartAst & 30.99 & 0.33 & 1.03 & 0.82 \\
TinyChart@768 & \textbf{33.35} & 1.53 & 1.64 & 1.89 \\ \bottomrule
\end{tabular}
\end{table}

\begin{table*}[ht!]
\caption{Ablation study. We train the models only using benchmark datasets in this experiment.}
\label{tab:ablation}
\scalebox{1.0}{
\begin{tabular}{@{}cccccccccccccc@{}}
\toprule
\multirow{2}{*}{Row} & \multirow{2}{*}{Resolution} & \multirow{2}{*}{\begin{tabular}[c]{@{}c@{}}GPT\\ PoT\end{tabular}} & \multirow{2}{*}{\begin{tabular}[c]{@{}c@{}}Template\\ PoT\end{tabular}} & \multirow{2}{*}{\begin{tabular}[c]{@{}c@{}}Visual Patch\\ Merge\end{tabular}} & \multirow{2}{*}{\begin{tabular}[c]{@{}c@{}}Visual\\ Length\end{tabular}} & \multirow{2}{*}{\begin{tabular}[c]{@{}c@{}}Inference\\ Throughput\end{tabular}} & \multicolumn{3}{c}{ChartQA} & Chart2Text & Chart2Table \\ \cmidrule(l){8-12} 

 &  &  &  &  &  &  & Direct & PoT & Combine & BLEU4 & RMS\(_{F1}\) \\ \midrule
1 & 384\(\times\)384 &\(\times\)&\(\times\)&\(\times\)& 729 & 3.73 it/s & 70.72 & - & - & 17.10 & 85.80 \\
2 & 384\(\times\)384 & \(\times\) & \checkmark &\(\times\)& 729 & 3.73 it/s & 71.12 & 55.44 & 73.00 & 17.04 & 87.68 \\
3 & 384\(\times\)384 & \checkmark & \checkmark &\(\times\)& 729 & 3.73 it/s & 72.44 & 76.88 & 79.48 & 16.67 & 87.30 \\ \midrule
4 & 512\(\times\)512 & \checkmark & \checkmark & \(\times\)& 1,296 & 2.38 it/s & 74.08 & 79.64 & 81.72 & 17.32 & 89.76 \\
5 & 512\(\times\)512 & \checkmark & \checkmark & \(r\)=12 & 984 & 2.84 it/s & 73.24 & 77.72 & 80.52 & 16.54 & 88.26 \\
6 & 512\(\times\)512 & \checkmark & \checkmark & \(r\)=15 & 906 & 3.26 it/s & 72.52 & 78.60 & 80.04 & 16.96 & 88.01 \\
7 & 512\(\times\)512 & \checkmark & \checkmark & \(r\)=20 & 776 & 3.65 it/s & 73.36 & 78.84 & 80.76 & 16.57 & 87.81 \\ \midrule
8 & 768\(\times\)768 & \checkmark & \checkmark &\(\times\)& 2,916 & OOM & - & - & - & - & - \\
9 & 768\(\times\)768 & \checkmark & \checkmark & \(r\)=84 & 732& 3.14 it/s & 73.24 & 77.72 & 81.04 & 16.43 & 88.90 \\ \bottomrule
\end{tabular}
}
\end{table*}

\begin{figure*}
\includegraphics[width=1.0\linewidth]{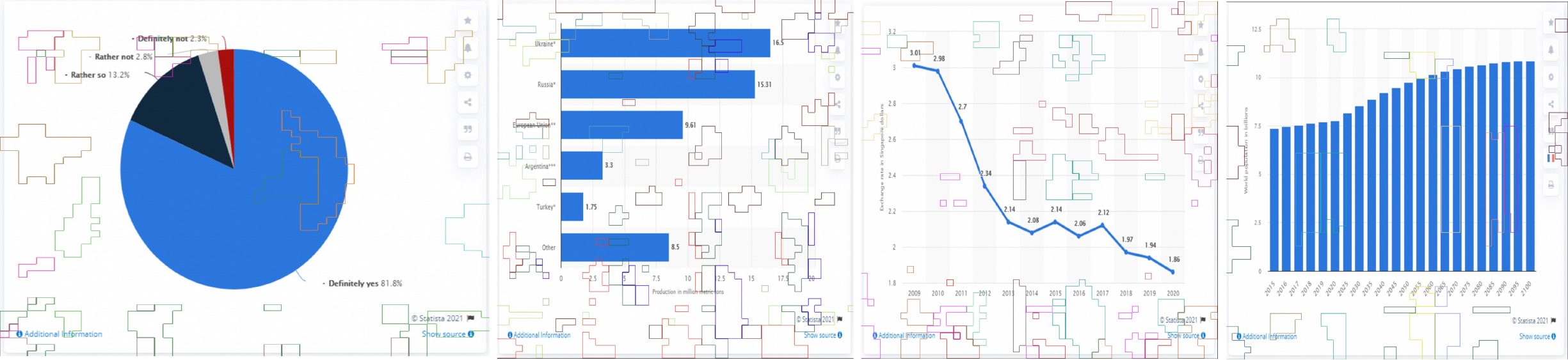}
\caption{Visual token merging visualization. Top 10 groups with the most merged tokens are outlined in different colors.}
\label{fig:vis_tokenmerge}
\end{figure*}

\subsection{Ablation Studies}
To verify the effectiveness of visual token merging and program-of-thoughts learning, we conduct ablation studies in Table~\ref{tab:ablation}. 

\noindentparagraph{\textbf{Ablation on PoT learning.}}The upper block in Table~\ref{tab:ablation} shows the performance of the model with and without the use of Program-of-Thoughts training data. Comparing Row 2 with Row 1, we observe that training solely with template-based PoT improves the model's ability to generate direct answers (71.12 vs. 70.72). This improvement is attributed to PoT learning enhances the model's reasoning abilities. At this point, the PoT answers produced by the model are less accurate than direct answers (55.44 vs. 71.12), which may be due to the inability of template-based PoT to cover all questions. However, when we ask the model to generate PoT answers for questions that require calculation and combine with direct answers, it outperforms solely direct answers (73.00 vs. 71.12). This indicates that PoT answers have advantages in computational problems. After incorporating GPT-based PoT into training, the performance of PoT answering surpasses direct answering (76.88 vs. 72.44), and both direct (72.44 vs. 71.12) and combined answering (79.48 vs. 73.00) show further improvements. These results confirm the effectiveness of our proposed Program-of-Thoughts learning method, suggesting that it not only strengthens the model's computational capabilities but also enhances overall problem-solving capability.

\noindentparagraph{\textbf{Ablation on Visual Token Merging.}}
The middle block in Table~\ref{tab:ablation} compares the performance with and without using visual token merging when the input resolution is 512\(\times\)512, and with different numbers of tokens to merge in each layer. Comparing Row 4 and Row 3, increasing the input resolution from 384 to 512 significantly improves the model's performance on three chart understanding benchmarks, demonstrating that high resolution is crucial for comprehending chart images. However, a direct increase in resolution leads to a substantial drop in the inference throughput (2.38 it/s vs. 3.73 it/s). The reason is that, given high-resolution images, the standard vision transformer produces a lengthy visual feature sequence that is then processed by the large language model. This brings considerable computational expenses. By adopting the visual token merging, we can control the length of the visual feature sequence by regulating the number of tokens to merge at each layer, and, thereby achieving efficient high-resolution encoding. When setting \(r\)=20, we attain an inference throughput nearly equal to that with an input resolution of 384\(\times\)384 (3.65 it/s vs. 3.73 it/s), while providing the performance benefits of higher resolutions.

\noindentparagraph{\textbf{Extending to higher resolution.}}
To further highlight the advantages of visual token merging, we increase the input resolution to 768 in the bottom block of Table~\ref{tab:ablation}. At this point, the length of the visual feature sequence is 2,916, which could not be trained using 32GB V100 due to insufficient VRAM. However, after employing the visual token merging module with \(r\)=84, the input sequence length is reduced to 732 and we can perform training normally. In this setting, the model's inference throughput is 3.14 it/s, and demonstrates a certain performance advantage in ChartQA (81.04 vs. 80.76) and Chart-to-Table (88.90 vs. 87.81). It illustrates that by utilizing visual token merging, we are able to leverage higher-resolution chart images under constrained resources, thereby improving performance.

\begin{figure*}
\includegraphics[width=0.975\linewidth]{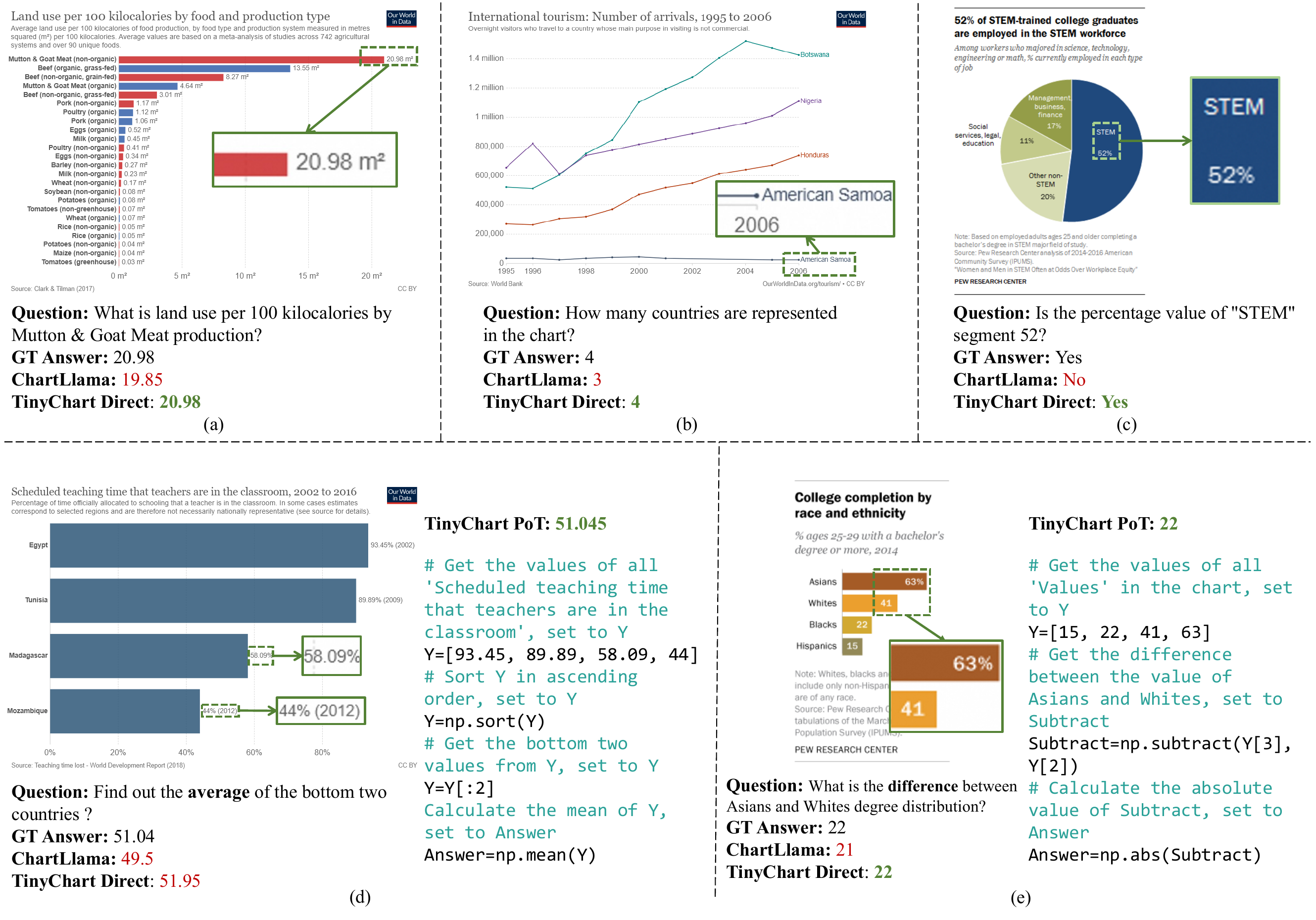}
\caption{Case studies on ChartQA. We compare TinyChart@768 with ChartLlama.}
\label{fig:vis_cases}
\end{figure*}

\begin{figure*}
\includegraphics[width=0.975\linewidth]{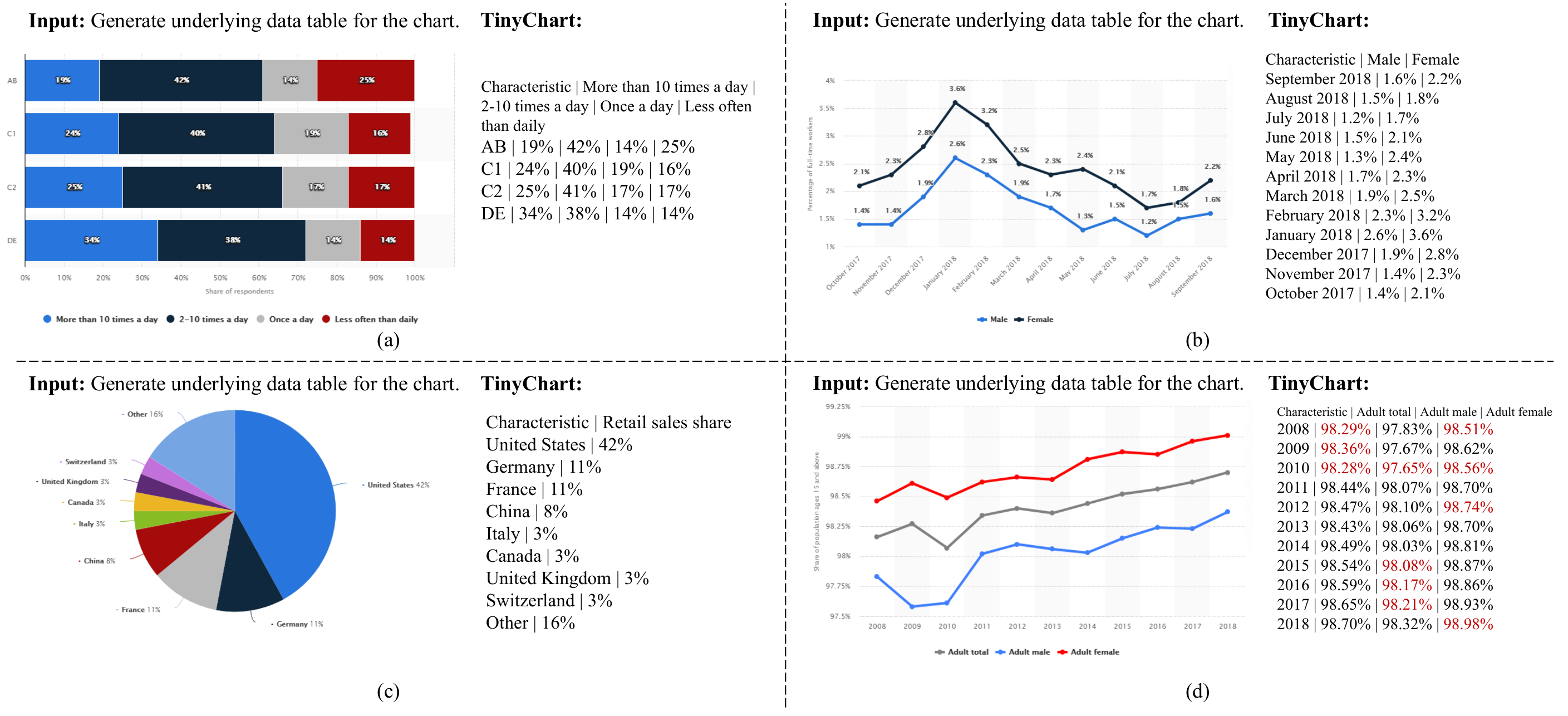}
\caption{Examples of chart-to-table extraction of TinyChart@768. The wrong values produced by the model are marked \textcolor[rgb]{0.686, 0.145, 0.098}{red}.}
\label{fig:table_cases}
\end{figure*}

\begin{figure*}
\includegraphics[width=1\linewidth]{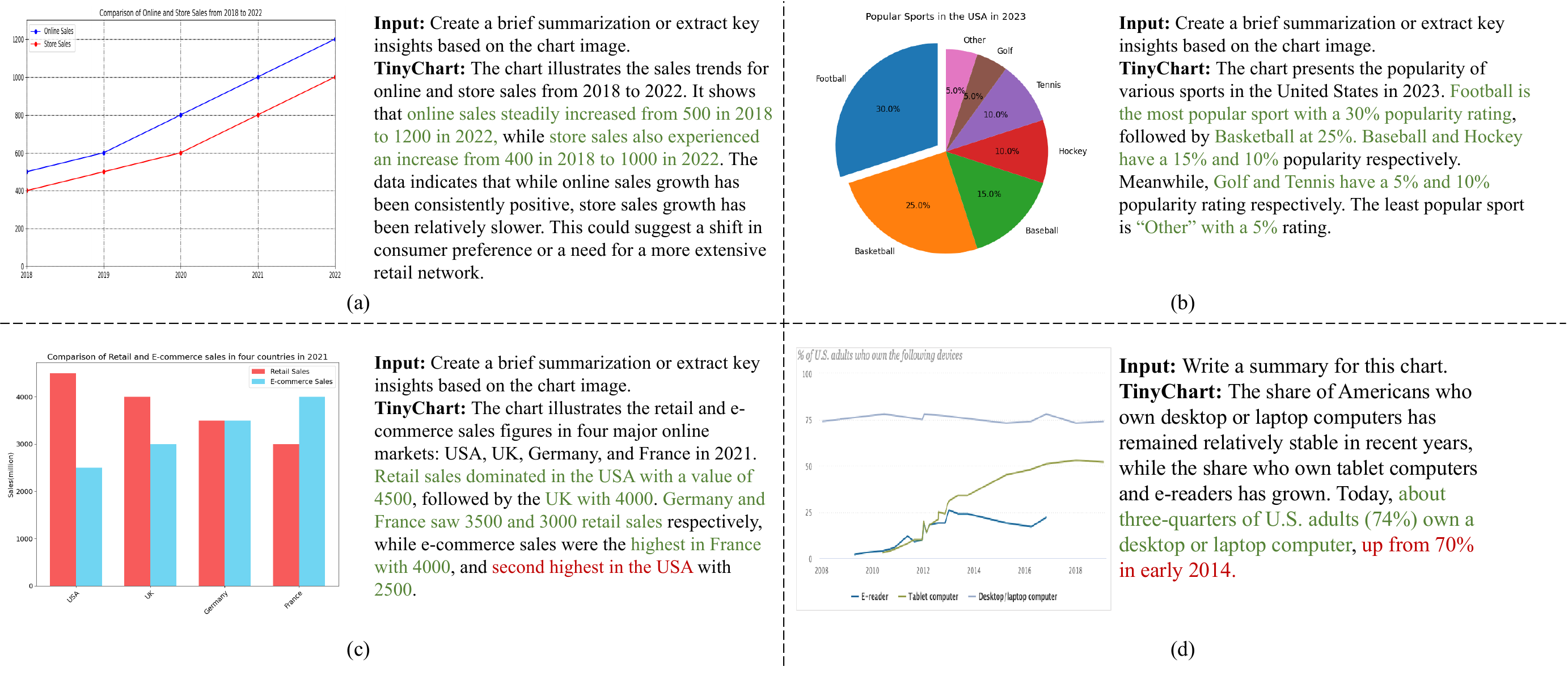}
\caption{Cases of chart-to-text generation by TinyChart@768. Correct contents are shown in \textcolor[rgb]{0.373, 0.506, 0.094}{green} and wrong contents are marked \textcolor[rgb]{0.686, 0.145, 0.098}{red}.}
\label{fig:summary_cases}
\end{figure*}

\begin{figure*}
\includegraphics[width=1\linewidth]{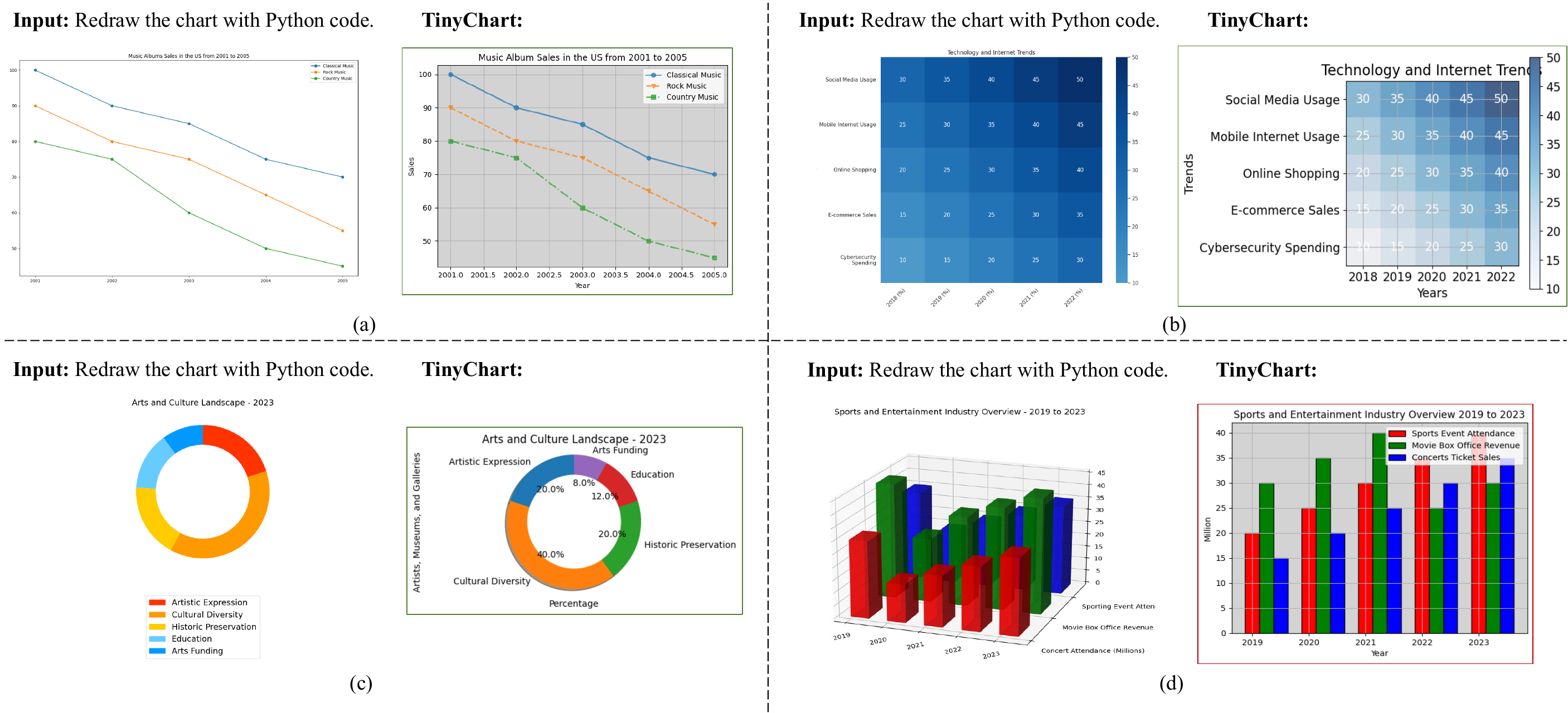}
\caption{Examples of chart redrawing. We present the resulting image after executing the Python code produced by the model. The bad case is with the \textcolor[rgb]{0.686, 0.145, 0.098}{red} bounding box.}
\label{fig:redraw_cases}
\end{figure*}

\subsection{Visualization}
To investigate the effects of visual token merging, we visualized the token merging results at the final layer of the vision transformer. In Figure~\ref{fig:vis_tokenmerge}, we visualize the top ten groups with the largest numbers of tokens. Each group is outlined with a different color. The visualization reveals that these largest groups typically correspond to blank or colored areas. By compressing these areas down to a single token for encoding, our visual token merging module can thus reduce the length of the encoded sequence without losing much information, thereby achieving efficient visual encoding.

\subsection{Case study}
We conduct case studies with TinyChart when conducting chart question answering, chart-to-table, chart-to-text, and chart redrawing in Figure~\ref{fig:vis_cases}, \ref{fig:table_cases}, \ref{fig:summary_cases}, and \ref{fig:redraw_cases}.

\noindentparagraph{\textbf{Chart Question Answering.}} In Figure~\ref{fig:vis_cases}, we present a case study on ChartQA. As shown in Figure~\ref{fig:vis_cases}~(a-c), much key information within the chart is provided by visually situated texts within the image, which requires the model to have the ability to process high-resolution images. Since ChartLlama only supports 336 resolutions, it struggles to retrieve accurate information in these charts. In contrast, thanks to the visual token merging, our TinyChart can accept higher-resolution inputs without introducing excessive computations. Thus it can successfully find clues related to the questions. Meanwhile, ChartLlama suffers from numerical errors when faced with calculative questions in Figure~\ref{fig:vis_cases}~(d-e), and our PoT (Program-of-Thoughts) learning method can accurately solve these problems. These examples further illustrate the advantages of our methods.
\noindentparagraph{\textbf{Chart-to-Table.}} 
For chart-to-table extraction, we find that our TinyChart model can successfully extractive values from several visually diverse charts in Figure~\ref{fig:table_cases}~(a-c), thanks to its excellent text recognition ability with high-resolution input. However, as shown in Figure~\ref{fig:table_cases}~(d), the model struggles to estimate the values of data points in the absence of OCR words. It seems that the model could make reasonable predictions based on surrounding points, but hardly provide accurate values based on the coordinate axis. This indicates that the model still lacks the ability to understand spatial relationships across large areas.
\noindentparagraph{\textbf{Chart-to-Text.}} 
From Figure~\ref{fig:summary_cases}, we observe that the model can understand the data presented in the chart and generate descriptions and summaries in natural language. Though it can retrieve the data values correctly, we find it sometimes produces contents that do match the chart as shown in Figure~\ref{fig:summary_cases}~(c-d). This may be due to the inherent limitations of hallucination in MLLMs~\cite{chair,pope,wang2023evaluation,amber}, and may be alleviated by addressing hallucinations~\cite{vcd,opera,jiang2024hallucination,less_eos}.
\noindentparagraph{\textbf{Chart redrawing.}} 
We present four cases of chart redrawing in Figure~\ref{fig:redraw_cases}. As shown in Figure~\ref{fig:redraw_cases}~(a-c), our TinyChart model can generate Python code to redraw visually diverse chart types including lines, heatmaps, and rings. However, it can be hard to draw unseen chart types such as 3D bar charts (Figure~\ref{fig:redraw_cases}~(d)). This may be mitigated by improving the coverage of different chart types in training data through automatic data construction techniques~\cite{chartllama,chartx}.
\section{Conclusion}
This paper introduces TinyChart, a chart understanding Multimodal Large Language Model with 3 billion parameters. To address the inefficiency of lengthy visual token sequences with high-resolution images, TinyChart injects a visual token merging module that merges similar vision tokens together, thereby enabling efficient encoding of high-resolution images. To tackle the challenges of learning numerical computations, we propose a Program-of-Thoughts learning method that trains the model to generate Python programs to answer questions. Our TinyChart model achieves state-of-the-art (SOTA) performance on multiple chart understanding benchmarks, surpassing existing 13 billion parameter chart MLLMs, and outperforms closed-source models like GPT-4V on ChartQA. Extensive ablation studies confirm the effectiveness of our methods. Our code and model are released at \href{https://github.com/X-PLUG/mPLUG-DocOwl/tree/main/TinyChart}{https://github.com/X-PLUG/mPLUG-DocOwl/tree/main/TinyChart}.

\balance
\bibliographystyle{ACM-Reference-Format}
\bibliography{sample-base}

\begin{figure*}[h!]
\centering
\scalebox{1.0}{
    \begin{tcolorbox}[colback=gray!5!white,colframe=white!0!black,title=Instructions to \texttt{gpt-3.5-turbo}]
Please generate a list of \textbf{assignment statements} in Python to answer the question of a chart. You can only use the following operators in each statement: \texttt{<function\_list>}\footnote{function\_list=['len', 'all', 'any', 'index', 'np.sort', 'np.abs', 'np.add', 'np.argmax', 'np.argmin', 'np.diff', 'np.divide','np.greater', 'np.greater\_equal', 'np.less', 'np.max', 'np.mean', 'np.median', 'np.min', 'np.subtract', 'np.sum', 'np.count\_nonzero', 'np.where', '+', '-', '*', '/', '>', '<', '=']}. Do not use any circulation or if-branch. Do not include any unnecessary statement that is not used. The chart is presented by a data table with color information. Note that the colors are estimated and may not match the description in the question. You can choose the most possible data if necessary. You must provide a one-line comment before each assignment statement. The last variable must be Answer.
Here are some examples:\\
\textbf{Example Input \#1:}\\
Chart title: Long-term price index in food commodities, 1850-2015, World, 1934 \\
Chart type: Horizontal bar chart \\
Chart table: \\
| Food  |  Long-term price index in food commodities, 1850-2015, World, 1934 |\\
|:---------------------------------------|------------:|\\
| Lamb (color: steelblue)  |   103.7  |\\
| Corn (color: sienna)    |   103.13 |\\
| Barley (color: mediumvioletred)  |  102.46 |\\
| Rye (color: tomato)  |   87.37 |\\
| Beef (color: sienna)  |  85.27 |\\
| Wheat (color: slategray)  |  83.73 |\\
Question: What is the sum of the price index that is greater than 100? \\
Answer: 309.29 \\
\textbf{Example Output \#1:}\\
\# Get the values of all 'Long-term price index of each food', set to Y \\
Y=[103.7, 103.13, 102.46, 87.37, 85.27, 83.73] \\
\# Check whether Y is greater than 100, set to Greater \\
Greater=np.greater(Y,100) \\
\# Find the indices where Greater is True, set to Indices \\
Indices=np.where(Greater)[0] \\
\# Get the values at position Indices, set to Y \\
Y=np.array(Y)[Indices] \\
\# Calculate the sum of all elements in Y, set to Answer \\
Answer=np.sum(Y) \\ 
\textbf{Input:}
\texttt{<target\_input>}\\
\textbf{Output:}
\end{tcolorbox}
}
\caption{Instructions used for generating GPT-based PoT.}
\label{fig:gpt_prompt}
\end{figure*}

\appendix
\section{ChartQA-PoT Details}
\subsection{Dataset Statistic}
We build ChartQA-PoT based on the images and questions in the training split of ChartQA~\cite{chartqa}. ChartQA-PoT consists of two subsets: Template-based PoT and GPT-based PoT. We present the statistics over ChartQA-PoT in Table~\ref{tab:chartqa_pot}. We find that answers provided by \texttt{gpt-3.5-turbo} are longer than template-based PoT, since they cover more diverse scenarios. 

\begin{table}[h]
\caption{Statistic over ChartQA-PoT}
\label{tab:chartqa_pot}
\begin{tabular}{lrrr}
\toprule
Statistic & \multicolumn{1}{c}{\begin{tabular}[c]{@{}c@{}}Template\\ PoT\end{tabular}} & \multicolumn{1}{c}{\begin{tabular}[c]{@{}c@{}}GPT\\ PoT\end{tabular}} & \multicolumn{1}{c}{\begin{tabular}[c]{@{}c@{}}ChartQA\\ PoT\end{tabular}} \\ \midrule
Num. of samples & 119,281 & 21,303 & 140,584 \\
Num. of images & 17,498 & 15,521 & 18,133 \\
Avg. answer characters & 319.38 & 381.23 & 328.75 \\
Avg. answer tokens & 117.70 & 136.01 & 120.48 \\ \bottomrule
\end{tabular}
\end{table}

\begin{figure}[htbp]
  \centering
  \begin{subfigure}[t]{0.236\textwidth}
    \centering
    \includegraphics[width=\textwidth]{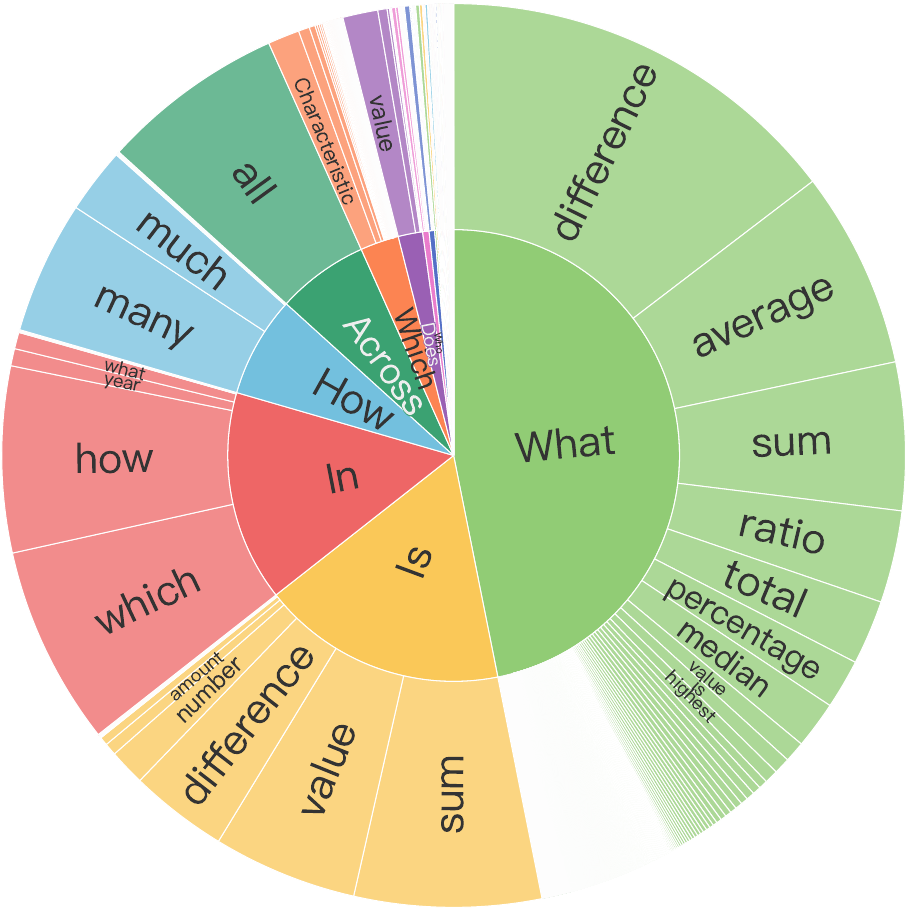}
    \caption{Template PoT.}
    \label{fig:temp_pot}
  \end{subfigure}
  \hfill
  \begin{subfigure}[t]{0.236\textwidth}
    \centering
    \includegraphics[width=\textwidth]{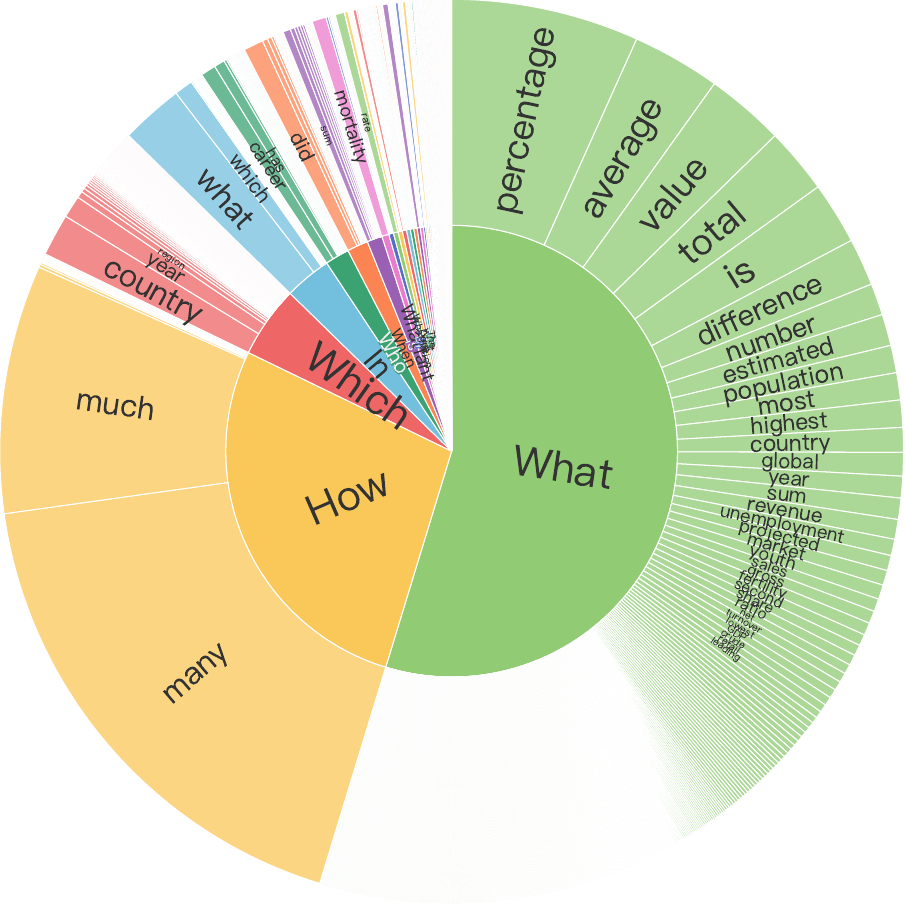}
    \caption{GPT PoT.}
    \label{fig:gpt_pot}
  \end{subfigure}

  \caption{First 2-gram of the questions in ChartQA-PoT after removing stop words.}
  \label{fig:gram_sun}
\end{figure}

We further present the first 2-gram words of the questions after removing stop words in Template-based PoT and GPT-based PoT in Figure~\ref{fig:gram_sun}. It is observed that GPT-PoT covers more diverse questions for `what' type questions, and questions in Template-based PoT are more evenly distributed across all question types.

\subsection{Instructions for GPT-based PoT}
Figure~\ref{fig:gpt_prompt} shows the instructions for constructing GPT-based PoT answers. Note that we prompt \texttt{gpt-3.5-turbo} to provide Python code consisting of assignment statements and avoid using loops or judgment statements. 
This can simplify the program and reduce syntax errors. We also provide meta information including the chart title, type, and colors to \texttt{gpt-3.5-turbo} since some questions rely on this information to answer.

\end{document}